# Emotion Detection in Reddit: Comparative Study of Machine Learning and Deep Learning Techniques

Maliheh Alaeddini

Syracuse University, NY

malaeddi@syr.edu

September 2024

## Abstract

Emotion detection is pivotal in human communication, as it significantly influences behavior, relationships, and decision-making processes. This study concentrates on text-based emotion detection by leveraging the GoEmotions dataset, which annotates Reddit comments with 27 distinct emotions. These emotions are subsequently mapped to Ekman's 6 basic categories: joy, anger, fear, sadness, disgust, and surprise. We employed a range of models for this task, including 6 machine learning models, 3 ensemble models, and Long Short-Term Memory (LSTM) model to determine the optimal model for emotion detection. Results indicate that the Stacking classifier outperforms other models in accuracy and performance. We also benchmark our models against EmoBERTa, a pre-trained emotion detection model, with our Stacking classifier proving more effective. Finally, the Stacking classifier is deployed via a Streamlit web application, underscoring its potential for real-world applications in text-based emotion analysis.

## Introduction

Emotions are complex, subjective experiences, often linked to psychological states such as mood, temperament, and personality. These experiences influence human behavior, impacting decision-making, reactions to stimuli, and interpersonal interactions. In the contemporary world, where mental health disorders such as stress, anxiety, and depression are increasingly prevalent, understanding emotions is more important than ever (Maruf et al., 2024). The automated detection of emotions in text has become a significant research focus, with applications spanning customer sentiment analysis, mental health support, and human-computer interaction (Canales & Martínez-Barco, 2014). Recognizing emotion from text is especially valuable in applications such as emotion-aware recommender systems, intelligent chatbots, and suicide prevention efforts. Emotion detection, however, differs significantly from sentiment analysis, which primarily categorizes text as positive or negative. Emotion detection instead requires identifying specific emotions, such as joy or sadness, a task which is inherently complex and often challenging, even for humans (Koufakou et al., 2022).

Emotion detection from text alone is difficult, as it lacks contextual and non-verbal cues such as tone or facial expressions that are often integral to human emotional understanding. The complexity of this task has motivated researchers to explore advanced machine learning and deep learning techniques that can improve classification accuracy and robustness (Al-Omari et al., 2020). This study focuses on Text-Based Emotion Detection (TBED), specifically using Reddit comments from the GoEmotions dataset, which provides labels for 27 emotions. These are mapped to Ekman's six primary emotions – joy, anger, fear, sadness, disgust, and surprise (Ekman, 1999). Our study applies a range of machine learning and deep learning techniques, including ensemble methods, to evaluate and identify the most effective model for emotion detection. By deploying our best-performing model via Streamlit, we aim to demonstrate its practical applications for real-world use.

This paper is organized as follows: Section II describes dataset preparation and preprocessing steps; Section III outlines model implementation and evaluation; Section IV presents a comparison with EmoBERTa; Section V discusses the deployment of our model; and Section VI concludes with limitations, challenges, and future directions.

## Dataset and Preprocessing

### Dataset Overview

The GoEmotions dataset, used in this study, comprises approximately 58,000 Reddit comments, each manually annotated for 27 emotion categories plus a neutral label. These labels encompass a broad spectrum of emotions, such as admiration, amusement, anger, annoyance, approval, caring, confusion, and curiosity, among others (Demszky et al., 2020). To simplify classification, the 27 original categories are mapped to Ekman's six basic emotions, along with a neutral category (Ekman, 1999). This mapping allows for focused, single-emotion classification, facilitating analysis across models. Following standard data preparation practices, the dataset was split into three segments – training, validation and test sets – providing ample data for model training and evaluation.

**Data Cleaning and Preprocessing Steps**

To streamline the dataset for analysis, extraneous columns were removed, and additional preprocessing steps were applied to standardize text inputs, especially for machine learning models:

***Handling Multiple Emotion Labels***: Some comments contain multiple emotion annotations, presenting a challenge for single-label classification. To manage this, a new column – "Class list" – was created by separating the annotations, and each comment was labeled with its most frequent emotion when multiple labels were present.

***Emotion Mapping***: Emotions were mapped to Ekman's basic categories, with a new "Mapped_emotion" column reflecting this simplified classification. Further, a "Single_emotion" column was created by selecting the dominant emotion for each comment, ensuring compatibility with single-label classification.

***Duplicate and Imbalance Handling***: Duplicate rows were removed to avoid redundancy, and the dataset's balance across emotion categories was analyzed.

***Text Normalization and Cleaning***: To optimize model performance across different approaches, two versions of the dataset were prepared:

- **Raw Dataset for Deep Learning Models:** Deep learning models, like LSTM, can extract complex, high-dimensional features directly from the raw text. Therefore, a minimally processed version of the dataset was preserved for these models, allowing them to leverage unaltered language patterns and contextual nuances that may enhance feature extraction.

- **Preprocessed Dataset for Machine Learning Models**: Traditional machine learning models often benefit from explicit feature engineering and data standardization. For these models, the data was normalized by converting all characters to lowercase and removing extraneous symbols. Emojis were converted to descriptive text, contractions were expanded, and abbreviations were standardized to reduce ambiguity (e.g., "lmao" to "laughing my ass off"). Additionally, repetitive letters were condensed, and punctuation was spaced for tokenization. Words were lemmatized to their base forms, and common stop words that do not contribute to emotion detection were removed, simplifying the dataset without sacrificing semantic meaning.

**Distribution of Emotions Across Data Splits**

The distribution of mapped emotions was visualized for each dataset split, as shown in Figure 1. This step identified significant class imbalances, with emotions like "joy" and "neutral" more prevalent than others.

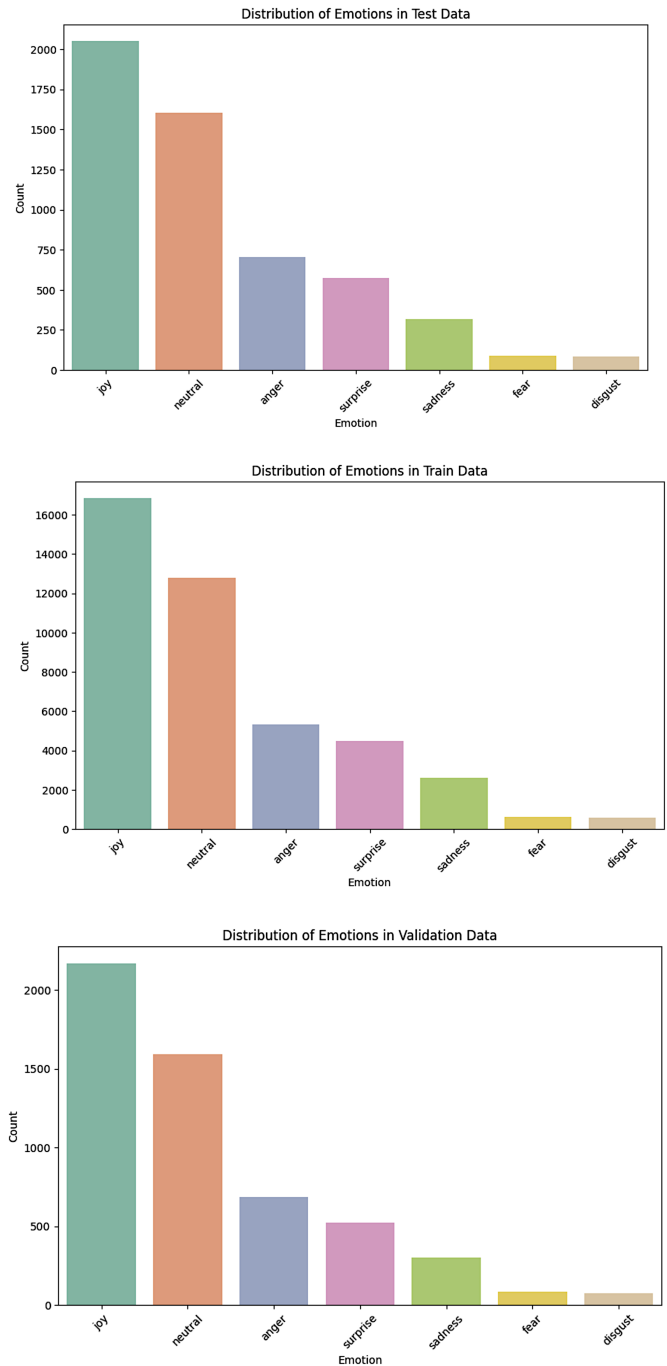

*Fig. 1. Distribution of Emotions Across Training, Test, and Validation Sets*

This dataset through its diverse range of emotional annotations and the preprocessing steps applied, offers a well-prepared foundation for implementing machine learning and deep learning models and comparative evaluation of these techniques in detecting emotions from Reddit comments.

## Model Implementation and Evaluation

**Tokenization and Feature Extraction**

To convert the text into numeric representations, we explored various tokenization techniques, including Term Frequency-Inverse Document Frequency (TF-IDF), Word2Vec, and GloVe embeddings. For machine learning models, TF-IDF was selected due to its effectiveness in transforming text data into structured features. TF-IDF captures term relevance by combining term frequency with inverse document frequency, assigning higher weights to important terms that appear frequently in a document but are rare across the dataset. This approach allows machine learning models to focus on meaningful features essential for emotion detection.

**Machine Learning Models**

For this study, we implemented a variety of supervised learning classification models to assess their effectiveness in TBED. Each model offers unique strengths, and detailed performance metrics were recorded to compare their efficacy in capturing complex emotional patterns:

***Decision Tree***: Decision Tree classifiers operate by splitting the data into branches based on decision rules, where each node represents a condition on an attribute, and each branch represents the outcome of the decision. The Decision Tree model achieved a moderate accuracy of 53%. While effective for classifying common emotions such as "joy" and "neutral" due to the clear patterns in these classes, it performed poorly on rarer emotions like "anger" and "fear," reflected in a low macro F1-score of 0.41. Decision Trees tend to overfit when applied to complex datasets with multiple classes, leading to limited generalization capabilities, particularly for less frequent emotions.

***Random Forest***: Random Forest is an ensemble method that generates multiple Decision Trees, each trained on different random subsets of the data. The model outputs the class that receives the most votes across these trees. By aggregating the results, Random Forest reduces the risk of overfitting and provides greater accuracy and stability. Our Random Forest model achieved an accuracy of 60%, performing best on common emotions like "joy" and "neutral." However, it struggled with classifying rarer emotions, resulting in lower recall scores. The model's macro F1-score of 0.42 and macro recall of 0.37 underscore its challenges with class imbalance, suggesting a need for further tuning or resampling techniques to improve performance on underrepresented classes.

*Support Vector Machine (SVM)*: SVM is a powerful model for high-dimensional data, creating a hyperplane that best separates data points from different classes. It works particularly well with small sample sizes and nonlinear classification tasks. Our SVM classifier achieved an accuracy of 63% and was able to capture common emotions such as "joy" and "neutral" effectively, as indicated by higher F1-scores for these classes. However, similar to other models, it had limited recall on rarer emotions like "fear" and "surprise," resulting in a macro average F1-score of 0.54. Although SVM performs well on frequently occurring classes, optimizing the model with balanced class representations may yield improvements across all emotion categories.

*Naive Bayes*: Naive Bayes is a probabilistic model based on Bayes' theorem, commonly applied to text classification due to its efficiency and simplicity. It assumes independence between predictors, which works well in some applications but can be a limitation in emotion detection. The Naive Bayes classifier's performance shows significant limitations in accurately identifying most emotions, with the model displaying high accuracy for only the "joy" and "neutral" classes, while severely underperforming on other emotions. The low recall across these underrepresented classes indicates that the model often fails to detect them, resulting in low F1-scores despite occasionally achieving high precision. This suggests that the model may struggle with class imbalance and the inherent complexity of emotional classification, as its assumption of feature independence makes it challenging to capture nuanced patterns in the data. Consequently, the model is heavily biased toward more frequent classes, limiting its overall effectiveness for this multi-class task.

*Logistic Regression*: Logistic Regression is a linear classifier that predicts the probability of a target variable, making it suitable for multi-class classification problems. The Logistic Regression classifier achieved an accuracy of 63% and performed particularly well on the "joy" class with an F1-score of 0.77. It also demonstrated strong performance on "neutral" and "sadness" classes, though recall for sadness was somewhat lower. For rarer classes like "anger" and "disgust," the model struggled to capture enough instances, with lower F1-scores between 0.26 and 0.46. The macro average F1-score of 0.50 reflects the need for further optimization, particularly by addressing class imbalance, to enhance its generalization to less frequent emotions.

***XGBoost*:** XGBoost is a gradient boosting algorithm that builds sequential models to improve upon the errors of previous models, making it effective for complex classification tasks. The XGBoost classifier achieved the highest accuracy among individual models at 64%. It performed exceptionally well on the "joy" class, achieving an F1-score of 0.77, and also displayed good recall for the "neutral" class. However, the model faced challenges with less frequent emotions such as "anger" and "fear" with lower precision and recall scores for these categories. The macro average F1-score of 0.53 reflects an imbalance in its ability to classify emotions, suggesting that tuning hyperparameters or addressing class imbalance could improve its detection of rarer emotions.

**Evaluation of Machine Learning Models**

After implementing and evaluating each machine learning model, a comparative analysis was conducted to identify the most effective models for emotion detection.

| Model | Accuracy | Precision | Recall | F1 Score |
|---|---|---|---|---|
| Logistic Regression | 0.63 | 0.64 | 0.63 | 0.61 |
| SVM | 0.63 | 0.62 | 0.63 | 0.62 |
| XGBoost | 0.61 | 0.64 | 0.61 | 0.60 |
| Random Forest | 0.60 | 0.63 | 0.60 | 0.56 |
| Decision Tree | 0.52 | 0.51 | 0.52 | 0.52 |
| Naive Bayes | 0.50 | 0.58 | 0.50 | 0.40 |

The performance of these models varied significantly across accuracy, precision, recall, and F1-score, offering insights into each model's strengths and limitations.

- *Accuracy*: Logistic Regression, SVM, and XGBoost emerged as the top-performing models in terms of accuracy, achieving scores of approximately 63% to 64%. These models demonstrated reliable classification for more common emotions, though accuracy decreased for rarer classes. Decision Tree and Naive Bayes, with lower accuracy scores of around 50%-53%, exhibited limitations in identifying less frequent emotions and struggled with the dataset's complexity.

- *Precision*: SVM and XGBoost showed the highest precision among the machine learning models, indicating a strong ability to classify positive instances correctly. The high precision of these models, especially on prevalent emotions, suggested that they could be valuable components in an ensemble learning framework, where minimizing false positives is essential.

- ***Recall***: Recall scores varied notably across models, with Logistic Regression, SVM, and XGBoost again leading in overall recall performance. However, recall scores were generally lower for rarer emotions across all models, highlighting the challenge of imbalanced classes in the dataset. Random Forest, with its ensemble nature, also performed moderately well in recall, capturing common patterns effectively.
- ***F1-Score***: Logistic Regression, SVM, and XGBoost achieved balanced F1-scores, reflecting a solid trade-off between precision and recall. The F1-scores were highest for these models, suggesting their robustness in handling diverse emotional classes and making them strong candidates for ensemble techniques. Naive Bayes, on the other hand, exhibited a low F1-score, indicating that its performance was heavily influenced by class imbalance.

Based on this evaluation, Logistic Regression, SVM, and XGBoost were selected as foundational models for ensemble learning due to their consistent performance across accuracy, precision, recall, and F1-score. Their balanced performance suggests that they can complement each other well in ensemble models, potentially improving overall classification accuracy and robustness in handling underrepresented emotions.

**Ensemble Models**

Ensemble learning is a robust approach in machine learning that combines multiple models to enhance overall performance, leveraging the strengths of each model while reducing individual weaknesses. For this study, we implemented three types of ensemble methods: Voting, Bagging, and Stacking. Each method demonstrated unique strengths in handling imbalanced classes and improving prediction accuracy across emotion categories.

***Voting Classifier***: The Voting Classifier combines predictions from multiple models to reach a final decision, using either hard or soft voting strategies. For this ensemble, we selected three high-performing models: XGBoost, Logistic Regression, and SVM. Both hard and soft voting methods were explored, with soft voting ultimately proving more effective due to its consideration of probability distributions across each model's predictions. The Voting Classifier achieved an accuracy of 62.7% with a weighted F1 score of 61% under soft voting conditions, showing improvement over individual models. This result indicates that the ensemble was able to capture a wider range of emotional nuances by balancing the probabilistic outputs of each base model, particularly boosting performance on common classes like "joy" and "neutral" while enhancing overall stability in predictions.

***Bagging Classifier***: Bagging, or Bootstrap Aggregating, is an ensemble technique where multiple instances of the same model are trained on different subsets of the dataset, and their predictions are aggregated. For this study, we implemented Bagging Classifiers using SVM, XGBoost, and Logistic Regression as base models, each trained with 10 estimators.

- **Bagging SVM:** This model achieved an accuracy of 60.2% and a weighted F1 score of 59%, showing resilience to overfitting and maintaining stable performance on frequent emotions. However, it still struggled with rarer emotions due to dataset imbalance.

- **Bagging XGBoost**: With an accuracy of 61.5% and a weighted F1 score of 60%, Bagging XGBoost showed the highest performance among the Bagging classifiers. The random sampling and aggregation helped in reducing overfitting and provided consistent predictions across emotion classes, though further improvements could be made for underrepresented emotions.

- **Bagging Logistic Regression**: This model achieved an accuracy of 61% and a weighted F1 score of 59%, demonstrating robust performance for frequent emotions but limited improvement on rare categories.

Bagging overall proved effective in enhancing the stability of each base classifier, with Bagging XGBoost performing most competitively, particularly when dealing with frequent emotions.

***Stacking Classifier*:** The Stacking Classifier, an advanced ensemble method, combines multiple base models and a meta-classifier, allowing the meta-classifier to learn from the predictions of the base models. For this implementation, we selected Random Forest, XGBoost, and SVM as base models, with Logistic Regression serving as the meta-classifier. The Stacking Classifier delivered the highest accuracy and F1 score among all models, outperforming individual models and other ensemble techniques. By combining the unique strengths of Random Forest, XGBoost, and SVM, the Stacking Classifier demonstrated superior robustness and adaptability in handling complex, multi-class emotion detection tasks. This approach allowed the meta-classifier to learn and correct errors made by the base models, resulting in an ensemble that excelled not only on common emotions like "joy" and "neutral" but also showed enhanced performance on rarer emotions.

The Stacking Classifier's success highlights the potential of ensemble learning to manage the complexities of multi-class emotion detection in text, as it effectively balanced predictions across frequent and less common emotion classes. Both the Voting and Stacking Classifiers showed competitive performance, with the Stacking Classifier achieving the best results, especially in accuracy and weighted F1 score. Bagging, particularly with XGBoost, also provided consistent performance, though it was slightly outperformed by the Voting and Stacking Classifiers in terms of overall accuracy.

**Deep Learning Model: LSTM**

Deep learning architectures, particularly LSTM networks, were employed to capture sequential dependencies in text data. LSTM networks are a type of Recurrent Neural Network (RNN) designed to overcome the vanishing gradient problem through a gating mechanism that selectively remembers or forgets information over long text sequences. By transforming text sequences through embedding and LSTM layers, the model captures complex language patterns relevant to emotion detection.

After tuning hyperparameters such as dropout rate, batch size, and unit size, the LSTM model achieved a weighted F1-score of 0.6. Although competitive, the LSTM did not outperform simpler machine learning models like Logistic Regression and SVM, suggesting that a deep learning model may not provide substantial advantages on this dataset.

**Overall Evaluation**

A comparison of the performance of different models for TBED provides some key insights. All models were evaluated based on accuracy, precision, recall, and F1-score, with higher values indicating better performance.

Among the machine learning models, Logistic Regression, SVM, and XGBoost emerged as top performers, each achieving around 63-64% accuracy. These models demonstrated balanced precision and recall, especially for common emotions like "joy" and "neutral", though they struggled with underrepresented classes. Naive Bayes and Decision Tree models were less effective, showing sensitivity to class imbalance and a tendency to overfit.

Ensemble methods showed clear improvements, leveraging multiple models to increase robustness. The Voting Classifier (accuracy of 62.7%) combined probabilistic outputs effectively but had limitations with rare emotions. The Bagging Classifier (with Bagging XGBoost achieving 61.5%) provided stability and handled frequent emotions well but offered limited gains on rare classes. Stacking outperformed all models, with its multi-model approach achieving the highest accuracy and F1-score by balancing strengths across both frequent and rare emotion classes.

The LSTM model provided competitive results (weighted F1-score of 0.6), but its complexity did not significantly outperform the top machine learning models, suggesting that deep learning may not always yield substantial advantages for TBED tasks with straightforward context.

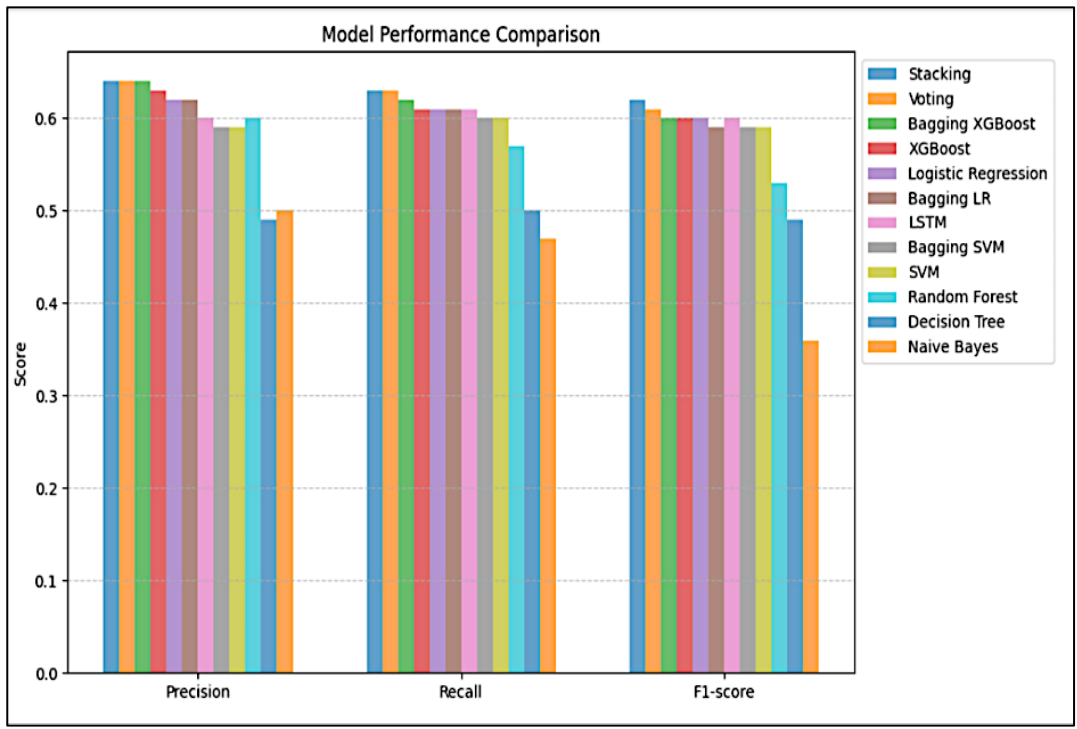

*Fig. 2. Comparative Performance of All Models*

Overall, the Stacking Classifier is the most effective approach, successfully managing class imbalances and capturing emotion nuances, making it ideal for complex, multi-class emotion detection tasks.

**Comparison with EmoBERTa**

To assess the effectiveness of our models, we conducted a performance comparison with EmoBERTa, a pre-trained model specifically designed for emotion detection. Built upon the RoBERTa (Robustly Optimized BERT Approach) architecture, EmoBERTa is trained on datasets like MELD and IEMOCAP, which incorporate text, audio, video, and dialogue data to capture nuanced emotional cues. We used the base version of EmoBERTa, which processes input sentences and outputs probabilities for each of Ekman's six basic emotions, plus a neutral category. For evaluation, we assigned each sentence the emotion label with the highest probability and compared these labels to the test set's true labels. This enabled direct measurement of EmoBERTa's accuracy and F1-score within our dataset's multi-class framework.

The EmoBERTa model achieved an accuracy of 50% and a weighted F1 score of 0.44 on our test set, lower than the Stacking Classifier, which had the highest overall performance among our implemented models. EmoBERTa's pre-training on diverse data sources, including non-text modalities, likely contributed to its generalizable but less focused performance on purely textual data from Reddit. While it captured common emotions like "joy" and "neutral" effectively, EmoBERTa showed limitations with rarer classes, similar to the individual machine learning models, due to its broader training objective.

The comparison highlights that while EmoBERTa offers robust generalizability for multi-modal emotion detection, specialized models tailored to the dataset's unique structure, such as the Stacking Classifier, may provide superior accuracy on single-modality, text-only tasks. The results suggest that pre-trained transformer models like EmoBERTa may benefit from fine-tuning on domain-specific datasets when applied to specialized emotion detection tasks, as they otherwise risk a trade-off in precision for domain-specific applications.

## Model Deployment Using Streamlit

To demonstrate the practical application of our emotion detection system, we deployed the best-performing model, the Stacking Classifier, using Streamlit, an open-source Python framework for creating interactive web applications tailored to data science and machine learning projects. This deployment allows users to interact with our model, testing its capabilities in real-time emotion detection on textual data.

**Application Architecture**

The application is designed with three primary components:

***Text Processing Module***: This module handles preprocessing for user-inputted text, ensuring that the data fed into the model aligns with the training data structure. This includes text normalization, tokenization, and other preprocessing steps used in the original training pipeline.

***Prediction Engine***: The core of the application, this engine implements the optimized Stacking Classifier to generate emotion predictions. Each prediction includes an associated probability score for Ekman's six basic emotions and a neutral category. The engine maps detected emotions to emojis, enhancing visual interpretation of results.

***Visualization Component***: This component provides users with a breakdown of prediction probabilities for each emotion, represented in a bar chart. By displaying probability scores, the application offers insight into the model's confidence for each predicted emotion, facilitating a more transparent user experience.

**User Interface and Functionality**

The Streamlit interface is designed for simplicity and accessibility. Users input text into a provided field and submit it for analysis.

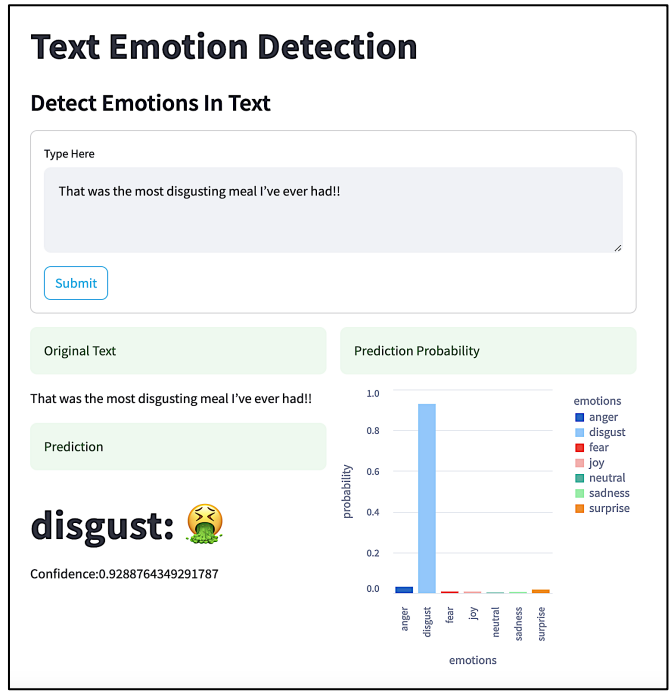

Upon submission, the interface displays the original text, the predicted emotion with an associated emoji, and a bar chart of probabilities for all emotion categories. To improve interactivity, the application uses a two-column layout:

- The left column displays the user's text, the predicted emotion, and an emoji, providing immediate feedback.
- The right column visualizes the probability scores for each possible emotion, helping users understand the model's confidence levels for each prediction.

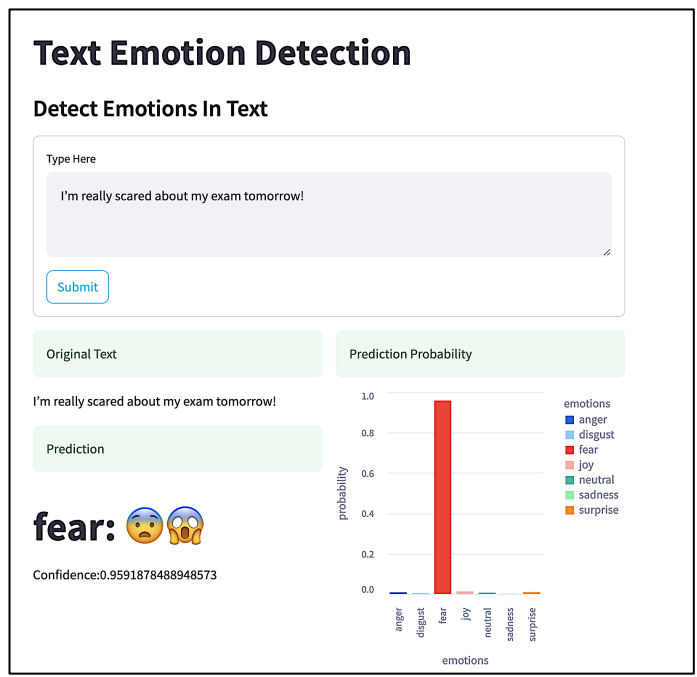

**Technical Specifications and Performance**

The application is optimized for scalability and quick response times. Streamlit's caching mechanisms ensure that the application operates smoothly under varying loads, maintaining response times around 1.8 seconds for typical text inputs. The modular design of the application allows for easy future updates, such as incorporating additional emotion categories or fine-tuning the model.

**User Experience and Practical Implications**

By deploying the Stacking Classifier through a user-friendly web application, our model becomes accessible to non-technical users, bridging the gap between research and real-world applications. The interactive interface, with visual enhancements like emojis and probability charts, aids in understanding nuanced emotional content, making it suitable for applications in fields like customer sentiment analysis, mental health support, and content moderation.

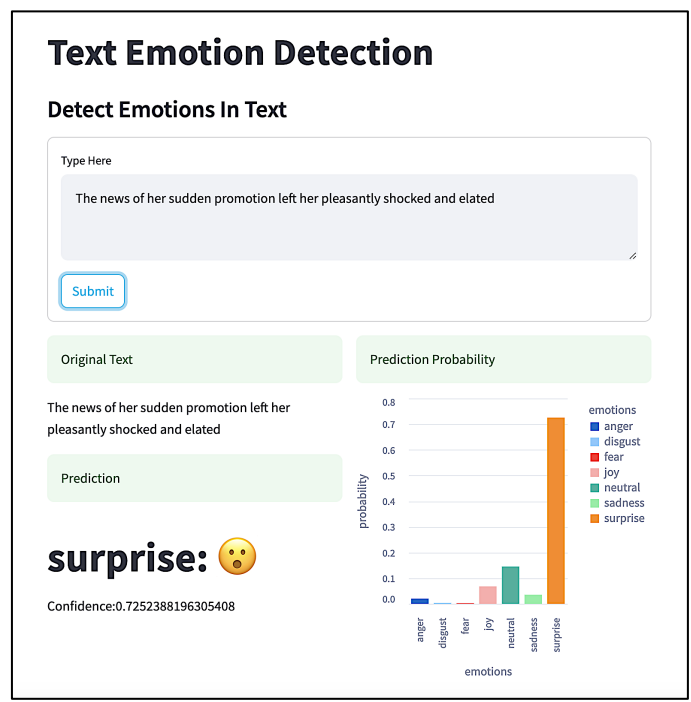

This deployment underscores the model's practical utility, offering a robust tool for emotion detection that can support a wide range of text-based emotion analysis tasks in real-time.

## Challenges, Conclusion, and Future Research Directions

**Challenges**

Throughout this project, we encountered several challenges inherent to emotion detection from text data. Key obstacles included:

***Class Imbalance***: The GoEmotions dataset exhibited significant class imbalances, with common emotions such as "joy" and "neutral" far more represented than other emotions like "fear" or "anger." This imbalance made it difficult for models to learn and generalize effectively for underrepresented classes, affecting recall scores in these categories.

***Complexity of Emotional Nuance in Text***: Emotions in language are inherently complex and can be challenging to interpret, especially without the context provided by tone or facial expressions. Text-based models are limited in capturing nuanced emotional expressions, and often struggled to interpret ambiguous language or detect mixed emotions accurately.

***Domain-Specific Language***: Social media platforms like Reddit use informal language, abbreviations, and slang, which can vary across communities. This domain-specific language can pose a challenge for emotion detection models trained on more formal datasets, as they may lack familiarity with colloquial expressions or slang.

***Interpretation of Sarcasm and Irony***: Sarcasm and irony are common in online interactions but are particularly challenging for TBED models. These expressions often convey emotions opposite to their literal meaning, requiring contextual understanding that text-only models lack. This limitation led to potential misclassifications in cases where sarcasm played a significant role in text interpretation.

***Computational Requirements***: Training complex models such as ensemble methods and deep learning networks demands significant computational power, which can be restrictive in research settings without high-performance infrastructure. This requirement limited the frequency and scale of model fine-tuning and experimentation.

***Generalizability Across Contexts***: Emotion detection models trained on Reddit data may face limitations when applied to other social media or text contexts, as language use, tone, and emotional expressions differ across platforms. Cross-platform generalization remains a challenge, potentially impacting the applicability of the model to broader TBED tasks.

**Conclusion**

This research underscores the effectiveness of combining machine learning, ensemble, and deep learning techniques for TBED using Reddit-based data from the GoEmotions dataset. Individual machine learning models such as Logistic Regression, SVM, and XGBoost demonstrated strong performance, but ensemble methods, particularly the Stacking Classifier, achieved superior accuracy and recall across emotion categories. The Stacking

Classifier's performance, surpassing even the EmoBERTa pre-trained model, highlights the advantage of custom-trained models for specific datasets.

The deployment of the Stacking Classifier through a Streamlit application demonstrates its potential real-world applications, making emotion detection accessible in real-time for users. This deployment showcases the practical utility of TBED models in fields like customer sentiment analysis, mental health monitoring, and content moderation, where real-time understanding of emotions can inform effective decisions and interventions.

**Future Research Directions**

This study opens avenues for further development and refinement of TBED models. Key areas for future exploration include:

*Addressing Class Imbalance*: Techniques such as data augmentation, stratified sampling, synthetic minority over-sampling (SMOTE), and acquiring additional data for underrepresented classes could improve model accuracy across all emotion categories, particularly for rare emotions.

*Multi-Modal Emotion Detection*: Integrating text with audio, visual, or contextual data may enhance emotion detection, providing models with richer context that can improve detection of emotions influenced by tone, body language, or situational factors.

*Improved Interpretation of Sarcasm and Irony*: Future research could explore methods to enhance model sensitivity to sarcasm and irony, perhaps through training on sarcasm-labeled datasets or incorporating sentiment polarity analysis to detect emotional subtext.

*Contextualized Transformer Models*: While this study focused on ensemble and LSTM models, future work could incorporate transformer-based models, which have shown promise in capturing complex context and semantics, potentially yielding better results for nuanced TBED tasks.

*Cross-Platform Adaptability*: Expanding the model's training data to include text from various platforms beyond Reddit could improve its generalizability, making it adaptable to different social media environments and types of digital communication.

In summary, this research highlights the potential of TBED models for understanding emotions in online interactions, with ensemble methods showing particular promise in capturing complex emotional expressions. Continued advancements in model architectures, data diversity, and computational efficiency will enhance the accuracy and applicability of TBED models, paving the way for systems that more accurately interpret and respond to human emotions in text.